# Distribution-restrained Softmax Loss for the Model Robustness

Hao Wang[1], Chen Li[2,3,4]*, Jinzhe Jiang[2,3,4], Xin Zhang[2,3,4], Yaqian Zhao[2,3,4], Weifeng Gong[2,5]

1. School of Computer and Artificial Intelligence, Zhengzhou University, Zhengzhou, China
2. Inspur Electronic Information Industry Co., Ltd, Jinan, China
3. State Key Laboratory of High-End Server & Storage Technology, Jinan, China
4. Shandong Hailiang Information Technology Institutes, Jinan, China
5. Zhengzhou Yunhai Information Technology Co., Ltd, Zhengzhou, China

* lichen03@inspur.com

## Abstract

Recently, the robustness of deep learning models has received widespread attention, and various methods for improving model robustness have been proposed, including adversarial training, model architecture modification, design of loss functions, certified defenses, and so on. However, the principle of the robustness to attacks is still not fully understood, also the related research is still not sufficient. Here, we have identified a significant factor that affects the robustness of models: the distribution characteristics of softmax values for non-real label samples. We found that the results after an attack are highly correlated with the distribution characteristics, and thus we proposed a loss function to suppress the distribution diversity of softmax. A large number of experiments have shown that our method can improve robustness without significant time consumption.

## 1. Introduction

While deep neural networks (DNNs) have demonstrated remarkable performance in a variety of applications including computer vision [1], speech recognition [2], and natural language processing [3], they are vulnerable to adversarial attacks, which involve the addition of small perturbations to input examples resulting in incorrect results with high confidence [4]-[8]. As DNNs are being widely used in various domains, ensuring their security by improving their robustness against adversarial attacks has become a critical research area.

Lots of defense techniques have been developed to enhance the adversarial robustness of DNNs [9]-[15]. A recent trend in adversarial defense is the use of certified defenses [16][17], which attempt to provide a guarantee that the model will not be fooled within an $l_p$ ball radius around original images. Nevertheless, this type of certified dense still have some limitations, such as high computational cost, lower robustness and higher requirements for data distribution. Adversarial training has been identified as the most effective approach [9][10]. However, adversarial training is time-consuming and resource-intensive, and sometimes even impairs the standard accuracy [18]. Based on adversarial training, many studies find that there is a trade-off between standard and robust accuracy [10],[18]-[20].

On the other hand, serval studies suggest that the analysis from adversarial training is not invariably accurate. Gillmer et al. [21] pointed out that, under the given setting, even small standard errors imply that most points can be proven to have misclassified points in their neighbouring region. In this setting, achieving perfect standard accuracy, which can be easily implemented with a simple classifier, is sufficient to achieve perfect adversarial robustness. Starting from Gaussian mixture model, Hu et al. [22] revealed two distinct effects: the first effect is a direct consequence of the constraint of adversarial robustness, which results in a degradation of the standard accuracy due to the optimizing direction change. The second effect is related to the class imbalance ratio between the two classes being considered, which leads to an increase in the difference of accuracy compared to standard training due to a reduction of "norm".

Hence, to overcome or relieve the standard accuracy problems, other methods have been proposed [11]-[15]. Goodfellow et al. [4] demonstrated that radial basis activation functions are more resistant to perturbations, but their deployment requires significant modifications to existing architectures. Papernot et al. [12] proposed a method to enhance the DNN perturbation robustness using distilled models. However, this method has some issues: (1) it requires dual training, which is costly, and (2) theoretically, the second model cannot be more accurate than the first model, which means that it will inevitably lead to some destroy of accuracy (although the test in the paper shows that the accuracy actually improved after distillation on CIFAR10, possibly because the baseline accuracy was not high, only 81.39%).

Wang et al. [23] proposed to use dropout during inference to introduce stochasticity and defend against adversarial attacks. On the other hand, Gu et al. [24] argued that the key issue of adversarial defense is to propose a suitable training process and objective function that can effectively enable the network to learn invariant regions around the training data. To this end, they proposed a deep contractive network to explicitly learn invariant features at each layer and restrict the change of dy with respect to dx by adding a term $\|dy/dx\|^2$ to the loss function, ensuring that the perturbations on x have little impact on y. They showed some promising preliminary results [24]. However, this penalty limits the ability of the deep contractive network compared to traditional DNNs [12].

It is worth noting that Rice et al. [25] argued that currently no method in isolation improves distinctly than early stopping. Additionally, Wu et al. [26] pointed out that early stopping can lead to a flatter weight loss landscape, which can result in a smaller robust generalization gap. But only if the training process is sufficiently, it can be beneficial to the test robustness.

In this paper, we propose a distribution-restrained softmax loss for adversarial robustness. Based on adversarial attack tests, we find that pre- and post-attack softmax are highly correlated. According to the relationship, we come up with a loss function to suppress the distribution diversity of softmax. By testing of this new loss function, we suggest that our method can improve the model robustness. While our research shares some similarities with previous studies [12][26], we believe that our loss function can serve as a significant complement to address different scenarios. Further comparative discussion will be conducted later.

## 2. Method

In this section, we introduce the method of iterative optimization to generate a transformation robust visualization images. In general, the input image is transformed by a certain operation, and the optimization is performed on this transformed image. Following, the transformation invariance is tested on the concerned model. The process is carried out iteratively until the convergence condition is achieved.

Previous versions of optimization with the zero image produced less recognizable images, while our method can give more informational visualization results.

## 2.1 Adversarial Attack

The target for adversary is to find an adversarial example $x_i'$ that can fool DNNs to make incorrect predictions. To make it unconspicuous for human, $x_i'$ should not be far away from $x_i$, by $\|x_i' - x_i\|_p \leq \epsilon$. There are many types of attacks have been proposed [27]-[29]. Here, we use Projected Gradient Descent (PGD) method as a probe to verify our method.

**Fast Gradient Sign Method (FGSM)**. FGSM perturbs the natural example $x_i$ by the step size of $\epsilon$ along the gradient direction:

$$x_i' = x_i + \epsilon \cdot sign(\nabla_{x_i} loss(f(x_i), y_i)) \qquad (1)$$

where f is the function of DNN model.

**I-FGSM**. Also, FGSM can be extended to an iterative version [30], which has been proposed by Kurakin et al.:

$$x_i^{N+1} = Clip_{x,\epsilon}\{x_i^N + \alpha \cdot sign\left(\nabla_{x_i} loss(f(x_i), y_i)\right)\} \qquad (2)$$

where N is the N-th iteration, $Clip_{x,\epsilon}$ indicates the attacked image is clipped within the $\epsilon$-ball of the last step, $\alpha$ is the value of perturbation.

**Projected Gradient Descent (PGD)**. PGD is an iterative method that perturbs the natural example $x_i$ by the certain value of $\eta$, and after each step of perturbation, it projects the adversarial example back to the adjacent of $x_i$:

$$x_i'^{(k+1)} = \prod_\epsilon(x_i'^{(k)} + \eta \cdot sign(\nabla_{x_i'} L(f\left(x_i'^{(k)}\right), y_i))) \qquad (3)$$

where L is the loss function, $\prod_\epsilon(\cdot)$ f is the projection operation, and $x_i'^{(k)}$ is the k-th step of the adversarial attack.

## 2.2 Distribution-restrained Softmax Loss Function (DRSL)

It is common to use cross entropy as a loss function for DNN:

$$L(f(x;\theta), y) = -1_y^T \log(softmax(f(x;\theta))) \tag{3}$$

where $\theta$ is the set of parameters of the classifier, $f(\cdot)$ is the function defined by DNN, and $1_y$ denotes the one-hot encoding of y.

However, there are some studies indicate that the softmax cross entropy doesn't guarantee a good robustness [26],[31],[37]-[40]. Following these works, we investigate the pre- and post-attack softmax probabilities on the MNIST dataset [35]. Figure 1 shows the softmax probabilities on the MNIST dataset. Here, we select the softmax probabilities after attack and the second maximum softmax probabilities before attack to illustrate. It is obvious to identify the pattern that they are roughly similar (except for the true label 1), which implies after attacks, the second maximum softmax probabilities trends to be the largest one.

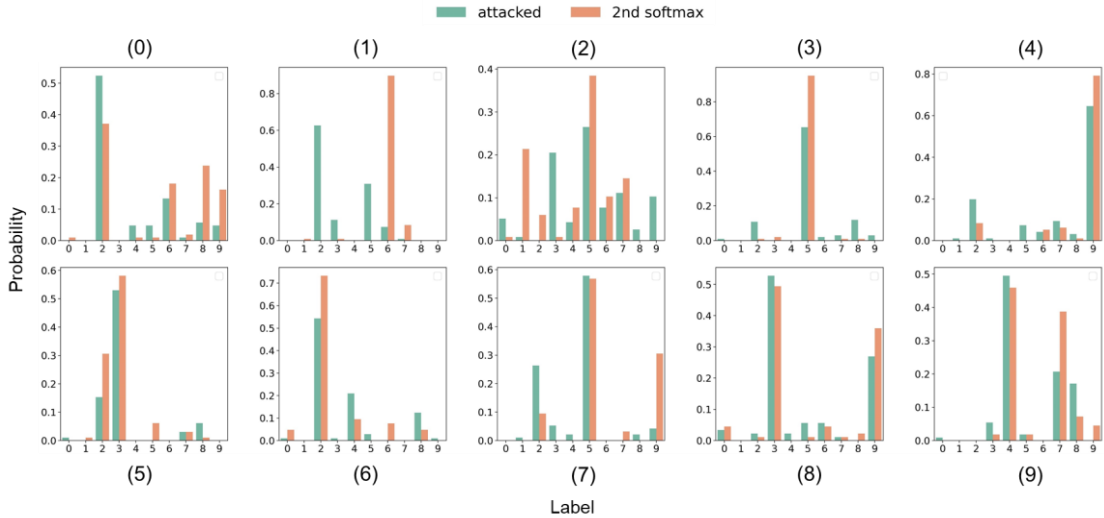

Figure 1. The softmax probabilities after attack and the second maximum softmax probabilities before attack on the MNIST dataset. (0)-(10) Corresponds to true labels of 0-10. The probability after attack is shown in green bar, and the second maximum softmax probabilities before attack is shown in red bar.

As we discussed before, trade-off between standard and robust accuracy is not a universal rule, especially for non- adversarial training method [21]-[24]. Gu et al. [24]

argued that the sensitivity of neural networks to adversarial examples is more related to inherent flaws in the training process and objective function, rather than the model architecture.

To further consolidate our findings, a standard accuracy-softmax distribution experiment is performed. Here, we adopt distance as a metric for stochastic of softmax distribution. Two common methods are used.

**Euclidean Distance**. Euclidean Distance is a well-used distance measure in the multi-dimensions of space. It is used to compare the absolute distances between two points in the dimensions of space:

$$d_E = \sqrt{\sum_{i=1}^{n}(a_i - b_i)^2} \tag{4}$$

where *a* and *b* are both n dimensional vectors.

**Cosine Distance**. Cosine Distance [41] computes the L2-normalized dot product of vectors. That is, if *a* and *b* are row vectors, their cosine similarity is defined as:

$$d_C(a, b) = \frac{ab^T}{\|a\|\|b\|} \tag{5}$$

Euclidean (L2) normalization projects the vectors onto the unit sphere, and their dot product is then the cosine of the angle between the points denoted by the vectors.

In the metric for stochasticity of softmax distribution case, the vector *a* and *b* are the real distribution of the output softmax of models and the ideal distribution, respectively. Here, we define the ideal case as a totally average distribution (For example, if there is a four-category classification, the ideal average distribution of the output softmax should be [0.25, 0.25, 0.25, 0.25]).

For a comprehensively evaluate, 3 architectures of DNN are tested: regularized Convolutional neural network VGG [32], multi-head attention based ViT [33], and a de-structural MLPMixer [42]. All the models are established in a comparable size, which will be discussed in the Section 3. Figure 2 gives the distance metric for stochasticity of softmax distribution. It shows our method rise the stochasticity of softmax distribution exactly, both on VGG and ViT models. Also, the result reveals that there are no significant positive correlations between standard accuracy and softmax distribution, that is, with the model accuracy increase, the softmax distribution doesn't become more sharpness. For cases of VGG, it shows even a negative correlation.

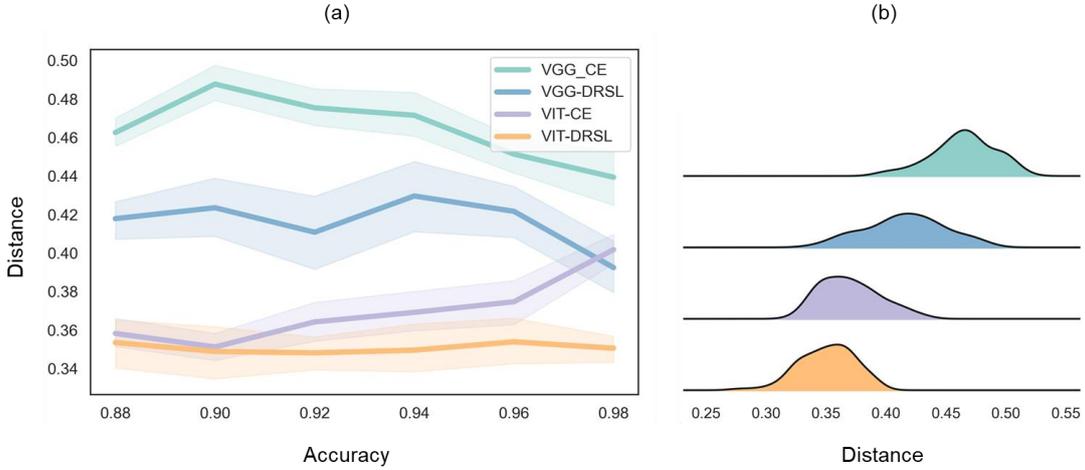

Figure 2. The distance metric for stochasticity of softmax distribution. (a) is the correlation between standard accuracy and the distance metric of different DNN models, (b) is the distance distribution of different models. CE is the abbreviation of standard cross entropy softmax, DRSL is our method.

Inspired by the preliminary result above, we assume the softmax distribution is a key factor of adversarial robustness, which is perhaps orthogonal to the standard optimizing direction and will not harm the accuracy significantly. Here, we propose the distribution-restrained softmax loss function towards robust DNN models:

$$L(f(x;\theta),y) = -1_y^T \log(softmax(f(x;\theta)))$$
$$+ \tau \cdot d(softmax(f(x;\theta)), avg) \qquad (3)$$

where $d(\cdot,\cdot)$ is the distance function, $\tau$ is the weight of the distance, $avg$ is the average distribution.

## 3. Experiments

In this section, we show the analysis of softmax distribution. Then cases of the robustness results are shown. Different loss functions are compared by adversarial robustness in Section 3.2, respect to different models and datasets. Also, the random noise robustness is tested in Section 3.3.

All networks used ReLUs in the hidden layers and softmax layers at the output. All reported experiments were repeated five times with random initialization of neural network parameters. We compared the proposed functions with Cross Entropy loss (CE), Generalized Cross Entropy loss (GCE) [31] and ours. All experiments were conducted with identical optimization procedures and architectures, changing only the loss functions. All the parameters size are restricted around 1.6 M. And the accuracy deviation of models are restricted to 0.5% (most cases are less than 0.1%). We conducted experiments using VGG [32] and ViT [33] models optimized with the default setting of Adam [34] on the MNIST [35] and CIFAR-10 [36] datasets.

## 3.1 Adversarial Robustness with Different Loss Functions

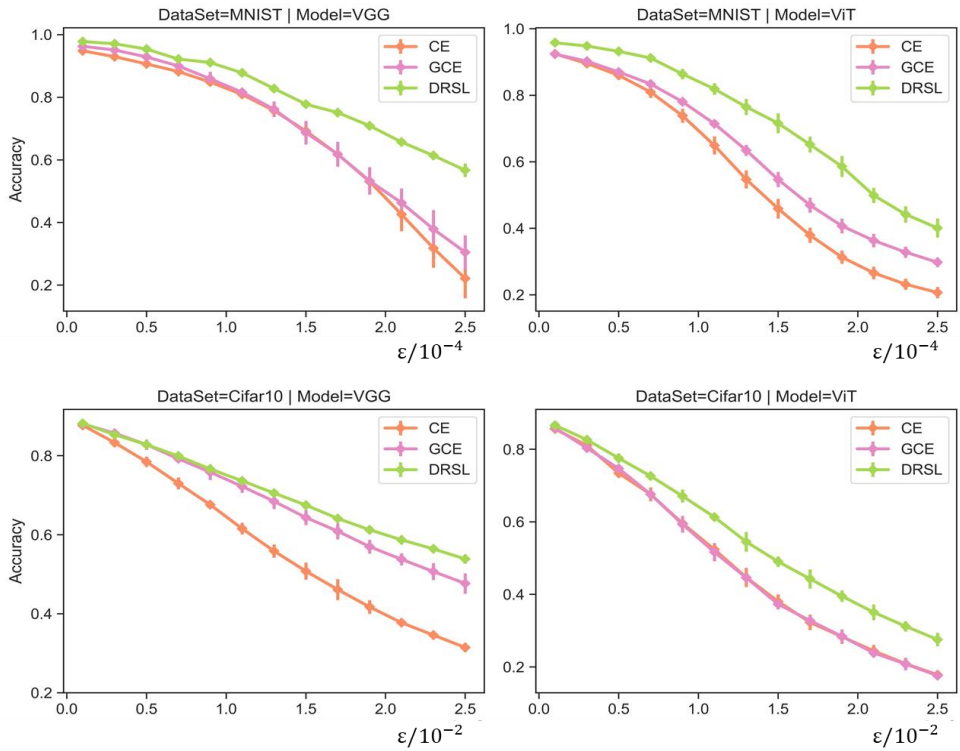

Figure 3. Adversarial Robustness with Different Loss Functions for different models. First row is for the MNIST dataset and the second is for CIFAR-10. And first column is VGG and the second is ViT model.

Experimental results of adversarial robustness with different loss functions are shown in figure 3. It reveals that the Distribution-restrained Softmax Loss outperform other loss functions in the VGG model. Besides MNIST dataset, CIFAR-10 are also used to train the models. In comparison, we also tested the loss functions for the ViT model. It can be observed that the advantage holds in the well-used ViT model.

## 3.2 The Label Noise Robustness with Different Loss Functions

Although our idea is based on analysis of adversarial perturbation, we are also curious about the effect of the noise. That's because the perturbation can be regard as a kind of noise, many works have pay close attention to their relation [43]-[45]. In this part, we will show a case study of label noise robustness with different loss functions [46]. To visually view the effect of loss function, output dimensions of DNNs are reduced from 10 to 2. Here, several dimensionality reduction method are used [47][48]. Figure 4 shows the result in MNIST dataset by t-SNE method, and others are shown in SI.

As shown in Figure 4(c), with the noise intensity increase, the test accuracy of all the models decrease. While the DRSL maintains the accuracy better than others. To make a further understanding of this phenomenon, dimensionality reduction methods are applied to visualize the softmax distribution. As shown in Figure 4(d), we can find that after a dimensionality reduction, DRSL gives a distinctive structure on the softmax distribution compared to others. Difference to clusters of CE and GCE that are "caterpillar-like", DRSL gives "nematode-like" clusters. After attacks, adversarial examples of other two models are distributed in the reduced space, while DRSL's are concentrated at the tip of clusters (Shown in Figure 4(d), second row. Adversarial examples are shown as black dots). That makes our method promising to discriminate adversarial examples, which need a further study in future. And under a label noise disturbed, models with other loss functions show a scene of chaos, while DRSL maintain a curve segment shape in the reduced space, which is relatively easy to distinguish of different categories. Therefore, this analysis shows that the DRSL function is not only helpful to adversarial robustness, but can also capture the label noise robustness.

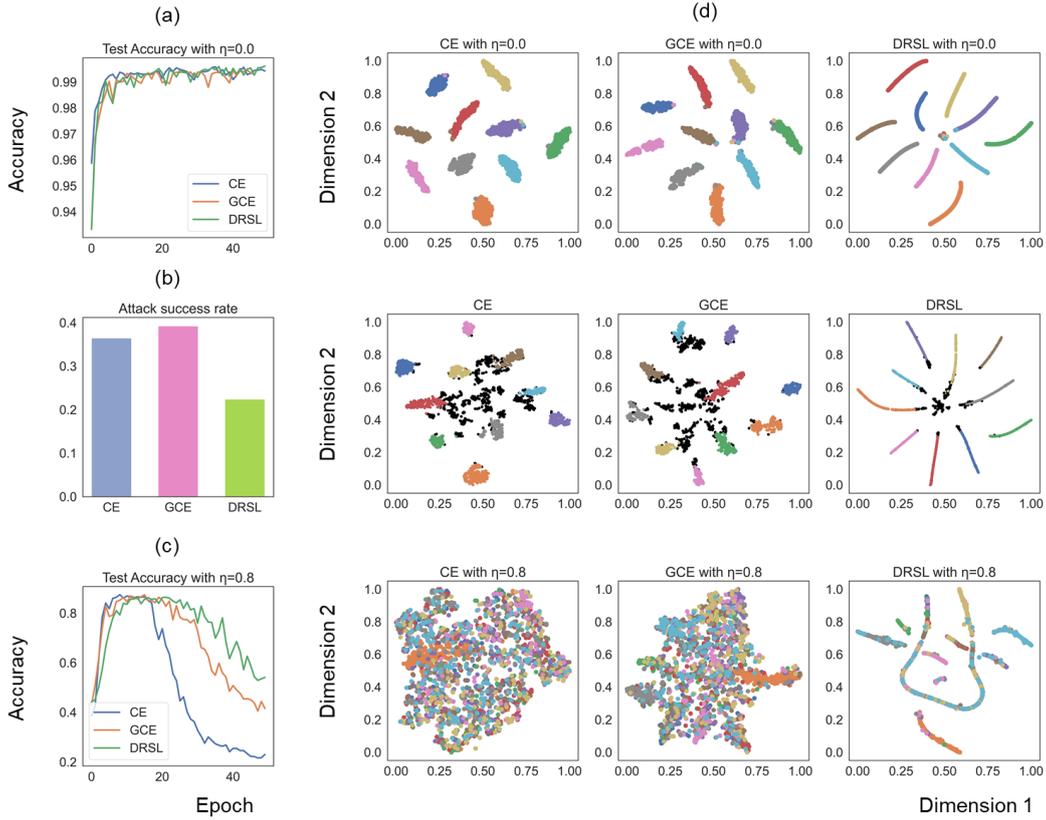

Figure 4. Dimensionality reduction perspective of different models. The first row is the classification cluster under dimensionality reduction after a standard training, the second row shows the result after attack of different models, and the third row shows the result after the label noise added. (a) is test accuracy with standard training, (b) is attack success rate, and (c) is the test accuracy with label noise η=0.8, respectively. Labels of category are shown in corresponding colors. And for the case of adversarial example, attacked data points are shown in black.

## 4. Conclusions

Robustness is a vital property for DNN models. In this paper, we have identified a significant factor that affects the robustness of models: the distribution characteristics of softmax values for non-real label samples. We found that the results after an attack are highly correlated with the distribution characteristics. And after the distribution diversity of softmax is suppressed in loss function, we find a significant improvement of model robustness. Although there are already some fundamental ideas, we believe that our loss function can serve as a significant complement to address different scenarios. This inspired us to further understand and improve models.

# Reference

[1] Kaiming He, Xiangyu Zhang, Shaoqing Ren, and Jian Sun. Deep residual learning for image recognition. In CVPR, 2016.

[2] Yisen Wang, Xuejiao Deng, Songbai Pu, and Zhiheng Huang. Residual convolutional ctc networks for automatic speech recognition. arXiv preprint arXiv:1702.07793, 2017.

[3] Ashish Vaswani, Noam Shazeer, Niki Parmar, Jakob Uszkoreit, Llion Jones, Aidan N. Gomez, Lukasz Kaiser, Illia Polosukhin. Attention is all you need. arXiv preprint arXiv:1706.03762, 2017.

[4] Ian J Goodfellow, Jonathon Shlens, and Christian Szegedy. Explaining and harnessing adversarial examples. In ICLR, 2015.

[5] A. Kurakin, I. Goodfellow, S. Bengio, Adversarial examples in the physical world, arXiv preprint arXiv:1607.02533, 2016.

[6] S. Komkov and A. Petiushko, AdvHat: Real-World Adversarial Attack on ArcFace Face ID System, 2020 25th International Conference on Pattern Recognition (ICPR), 2021, pp. 819-826

[7] I. Evtimov, K. Eykholt, E. Fernandes, T. Kohno, B. Li, A. Prakash, A. Rahmati, D. Song, Robust Physical-World Attacks on Deep Learning Models, arXiv preprint arXiv:1707.08945, 2017

[8] N. Morgulis, A. Kreines, S. Mendelowitz, Y. Weisglass, Fooling a Real Car with Adversarial Traffic Signs, arXiv preprint arXiv:1907.00374, 2019

[9] Aleksander Madry, Aleksandar Makelov, Ludwig Schmidt, Dimitris Tsipras, and Adrian Vladu. Towards deep learning models resistant to adversarial attacks. In International Conference on Learning Representations (ICLR), 2018.

[10] Hongyang Zhang, Yaodong Yu, Jiantao Jiao, Eric P Xing, Laurent El Ghaoui, and Michael I Jordan. Theoretically principled trade-off between robustness and accuracy. In International Conference on Machine Learning (ICML), 2019b.

[11]  Shixiang Gu, Luca Rigazio. Toward deep neural network architectures robust to adversarial examples. In International Conference on Learning Representations (ICLR), 2015.

[12]  Nicolas Papernot, Patrick McDaniel, Xi Wu, Somesh Jha, and Ananthram Swami. Distillation as a Defense to Adversarial Perturbations against Deep Neural Networks. arXiv preprint arXiv:1511.04508, 2015.

[13]  Siyue Wang, Xiao Wang, Pu Zhao, Wujie Wen, David Kaeli, Peter Chin, Xue Lin. Defensive Dropout for Hardening Deep Neural Networks under Adversarial Attacks. arXiv preprint arXiv:1809.05165,2018.

[14]  A. Raghunathan, J. Steinhardt, and P. Liang. Certified defenses against adversarial examples. ICLR, 2018.

[15]  E.Wong and Z. Kolter. Provable defenses against adversarial examples via the convex outer adversarial polytope. in International Conference on Machine Learning. PMLR, 2018.

[16]  A. Raghunathan, J. Steinhardt, and P. Liang. Certified defenses against adversarial examples. ICLR, 2018.

[17]  E.Wong and Z. Kolter. Provable defenses against adversarial examples via the convex outer adversarial polytope. In International Conference on Machine Learning. PMLR, 2018, pp. 5286–5295.

[18]  Dimitris Tsipras, Shibani Santurkar, Logan Engstrom, Alexander Turner, Aleksander Madry. Robustness May Be at Odds with Accuracy. arXiv preprint arXiv:1805.12152

[19]  Aditi Raghunathan, Sang Michael Xie, Fanny Yang, John C. Duchi, Percy Liang. Adversarial Training Can Hurt Generalization. arXiv preprint arXiv:1906.06032, 2019

[20] Dong Su, Huan Zhang, Hongge Chen, Jinfeng Yi, Pin-Yu Chen, and Yupeng Gao. Is Robustness the Cost of Accuracy? A Comprehensive Study on the Robustness of 18 Deep Image Classification Models. arXiv preprint arXiv:1808.01688, 2018.

[21] Justin Gilmer, Luke Metz, Fartash Faghri, Samuel S. Schoenholz, Maithra Raghu, Martin Wattenberg, & Ian Goodfellow. The Relationship Between

High-Dimensional Geometry and Adversarial Examples. arXiv preprint arXiv:1801.02774, 2018.

[22] Yuzheng Hu, FanWu, Hongyang Zhang, Han Zhao. Understanding the Impact of Adversarial Robustness on Accuracy Disparity. arXiv preprint arXiv:2211.15762, 2022.

[23] Siyue Wang, Xiao Wang, Pu Zhao, Wujie Wen, David Kaeli, Peter Chin, Xue Lin. Defensive Dropout for Hardening Deep Neural Networks under Adversarial Attacks. arXiv preprint arXiv:1809.05165, 2018.

[24] Shixiang Gu, Luca Rigazio. Towards Deep Neural Networks Architectures Robust to Adversarial Examples. In Proceedings of the 2015 International Conference on Learning Representations. Computational and Biological Learning Society, 2015.

[25] Leslie Rice, Eric Wong, J. Zico Kolter. Overfitting in adversarially robust deep learning. arXiv preprint arXiv:2002.11569, 2020.

[26] Dongxian Wu, Shu-Tao Xia, Yisen Wang. Adversarial Weight Perturbation Helps Robust Generalization. In 34th Conference on Neural Information Processing Systems, 2020.

[27] Ian J Goodfellow, Jonathon Shlens, and Christian Szegedy. Explaining and harnessing adversarial examples. In ICLR, 2015.

[28] Aleksander Madry, Aleksandar Makelov, Ludwig Schmidt, Dimitris Tsipras, and Adrian Vladu. Towards deep learning models resistant to adversarial attacks. In ICML, 2018.

[29] Nicholas Carlini and David Wagner. Towards evaluating the robustness of neural networks. In S&P, 2017.

[30] A. Kurakin, I. Goodfellow, and S. Bengio. Adversarial machine learning at scale. In International Conference on Learning Representations, 2017.

[31] Zhilu Zhang Mert R. Sabuncu. Generalized Cross Entropy Loss for Training Deep Neural Networks with Noisy Labels. In 32nd Conference on Neural Information Processing Systems, 2018.

[32] Simonyan K, Zisserman A. Very Deep Convolutional Networks for Large-Scale

Image Recognition. arXiv preprint arXiv:1409.1556, 2014.

[33] Alexey Dosovitskiy, Lucas Beyer, Alexander Kolesnikov, Dirk Weissenborn, Xiaohua Zhai, Thomas Unterthiner, Mostafa Dehghani, Matthias Minderer, Georg Heigold, Sylvain Gelly, Jakob Uszkoreit, Neil Houlsby. An Image is Worth 16x16 Words: Transformers for Image Recognition at Scale. arXiv preprint arXiv:2010.11929, 2020.

[34] Diederik P Kingma and Jimmy Ba. Adam: A method for stochastic optimization. arXiv preprint arXiv:1412.6980, 2014.

[35] LeCun, Yann and Cortes, Corinna and Burges, CJ. MNIST handwritten digit database. 2010.

[36] Alex Krizhevsky and Geoffrey Hinton. Learning multiple layers of features from tiny images. 2009.

[37] Tianyu Pang, Kun Xu, Yinpeng Dong, Chao Du, Ning Chen, Jun Zhu. Rethinking softmax cross-entropy loss for adversarial robustness. ICLR 2020.

[38] Weiyang Liu, Yandong Wen, Zhiding Yu, and Meng Yang. Large-margin softmax loss for convolutional neural networks. In International Conference on Machine Learning (ICML), 2016

[39] Hao Wang, Yitong Wang, Zheng Zhou, Xing Ji, Dihong Gong, Jingchao Zhou, Zhifeng Li, and Wei Liu. Cosface: Large margin cosine loss for deep face recognition. In Proceedings of the IEEE Conference on Computer Vision and Pattern Recognition (CVPR), pp. 5265–5274, 2018.

[40] Jiankang Deng, Jia Guo, Niannan Xue, and Stefanos Zafeiriou. Arcface: Additive angular margin loss for deep face recognition. In Proceedings of the IEEE Conference on Computer Vision and Pattern Recognition (CVPR), pp. 4690–4699, 2019.

[41] C.D. Manning, P. Raghavan and H. Schütze (2008). Introduction to Information Retrieval. Cambridge University Press. https://nlp.stanford.edu/IR-book/html/htmledition/the-vector-space-model-for-scoring-1.html

[42] Ilya Tolstikhin, Neil Houlsby, Alexander Kolesnikov, Lucas Beyer, Xiaohua Zhai, Thomas Unterthiner, Jessica Yung, Andreas Steiner, Daniel Keysers, Jakob Uszkoreit, Mario Lucic, Alexey Dosovitskiy. MLP-Mixer: An all-MLP

Architecture for Vision. arXiv preprint arXiv:2105.01601, 2021.

[43] Akshay Agarwal, Mayank Vatsa, Richa Singh, and Nalini K. Ratha. Noise is Inside Me! Generating Adversarial Perturbations with Noise Derived from Natural Filters. In Proceedings of the IEEE Conference on Computer Vision and Pattern Recognition Workshop (CVPR), 2020.

[44] Chengjun Tang, Kun Zhang, Chunfang Xing, Yong Ding, Zengmin Xu. Perlin Noise Improve Adversarial Robustness. arXiv preprint arXiv:2112.13408, 2021.

[45] Fei Wu, Wenxue Yang, Limin Xiao and Jinbin Zhu. Adaptive Wiener Filter and Natural Noise to Eliminate Adversarial Perturbation. Electronics 9, 1634, 2020.

[46] Aritra Ghosh, Himanshu Kumar, P. S. Sastry. Robust Loss Functions under Label Noise for Deep Neural Networks. arXiv preprint arXiv:1712.09482, 2017.

[47] Lever, J., Krzywinski, M. & Altman, N. Principal component analysis. Nat Methods 14, 641–642, 2017.

[48] Maaten, L.v.d. and Hinton, G. Journal of Machine Learning Research, 9, 2579-2605, 2008.